\documentclass{article}
\usepackage[final]{corl_2021}

\usepackage[printonlyused,nohyperlinks]{acronym}
\usepackage{subcaption}

\usepackage{algorithm}
\usepackage{algorithmicx}
\usepackage[noend]{algpseudocode}
\usepackage{etoolbox}
\usepackage{amssymb}
\usepackage{amsmath}
\usepackage{svg}
\usepackage{xspace}
\usepackage{booktabs}
\usepackage{float}

\usepackage{graphicx}
\usepackage{multirow}
\usepackage{verbatim}

\usepackage{array}
\usepackage{adjustbox}
\usepackage{wrapfig}

\usepackage{tikz}
\usetikzlibrary{shapes, arrows, positioning, fit, backgrounds}
\usetikzlibrary{calc}
\usetikzlibrary{decorations.pathmorphing}
\usepackage{caption}
\usepackage{siunitx}
\usepackage{hyperref}       %
\usepackage{url}  
\usepackage[T1]{fontenc}
\usepackage[utf8]{inputenc}
\newcommand*{\methodname}{ReLMM\xspace}

\tikzstyle{block} = [rectangle, draw, fill=blue!20, 
    text centered, rounded corners, minimum height=1em, node distance=1.5cm]
\tikzstyle{line} = [draw, -latex']
\tikzstyle{cloud} = [draw, rectangle,fill=red!20, node distance=1.5cm and 1cm,
    minimum height=1em]

\usepackage{array,multirow}
\usepackage{cprotect}
\usepackage{csquotes}

\algrenewcommand\algorithmicindent{0.5em}%

\usetikzlibrary{tikzmark}
\usetikzlibrary{calc}

\errorcontextlines\maxdimen

\newcommand{\ALGtikzmarkcolor}{black}%
\newcommand{\ALGtikzmarkextraindent}{2pt}%
\newcommand{\ALGtikzmarkverticaloffsetstart}{-.5ex}%
\newcommand{\ALGtikzmarkverticaloffsetend}{-.5ex}%
\makeatletter
\newcounter{ALG@tikzmark@tempcnta}

\newcommand\ALG@tikzmark@start{%
    \global\let\ALG@tikzmark@last\ALG@tikzmark@starttext%
    \expandafter\edef\csname ALG@tikzmark@\theALG@nested\endcsname{\theALG@tikzmark@tempcnta}%
    \tikzmark{ALG@tikzmark@start@\csname ALG@tikzmark@\theALG@nested\endcsname}%
    \addtocounter{ALG@tikzmark@tempcnta}{1}%
}

\def\ALG@tikzmark@starttext{start}
\newcommand\ALG@tikzmark@end{%
    \ifx\ALG@tikzmark@last\ALG@tikzmark@starttext
    \else
        \tikzmark{ALG@tikzmark@end@\csname ALG@tikzmark@\theALG@nested\endcsname}%
        \tikz[overlay,remember picture] \draw[\ALGtikzmarkcolor] let \p{S}=($(pic cs:ALG@tikzmark@start@\csname ALG@tikzmark@\theALG@nested\endcsname)+(\ALGtikzmarkextraindent,\ALGtikzmarkverticaloffsetstart)$), \p{E}=($(pic cs:ALG@tikzmark@end@\csname ALG@tikzmark@\theALG@nested\endcsname)+(\ALGtikzmarkextraindent,\ALGtikzmarkverticaloffsetend)$) in (\x{S},\y{S})--(\x{S},\y{E});%
    \fi
    \gdef\ALG@tikzmark@last{end}%
}

\apptocmd{\ALG@beginblock}{\ALG@tikzmark@start}{}{\errmessage{failed to patch}}
\pretocmd{\ALG@endblock}{\ALG@tikzmark@end}{}{\errmessage{failed to patch}}
\makeatother
\makeatletter
\def\thanks#1{\protected@xdef\@thanks{\@thanks
        \protect\footnotetext{#1}}}
\makeatother

\let\svthefootnote\thefootnote
\newcommand\blankfootnote[1]{%
	\let\thefootnote\relax\footnotetext{#1}%
	\let\thefootnote\svthefootnote%
}
\let\svfootnote\footnote
\renewcommand\footnote[2][?]{%
	\if\relax#1\relax%
	\blankfootnote{#2}%
	\else%
	\if?#1\svfootnote{#2}\else\svfootnote[#1]{#2}\fi%
	\fi
}

\title{
Fully Autonomous Real-World Reinforcement Learning with Applications to Mobile Manipulation
}

\author{{Charles Sun$^{*}$}, {Jędrzej Orbik$^{*}$}, {Coline Devin}, {Brian Yang},\\ \textbf{{Abhishek Gupta}, {Glen Berseth}, {Sergey Levine}} \\
{\{charlesjsun,jedrzej.orbik,coline,brianhyang,abhigupta,gberseth,svlevine\}@berkeley.edu}
\\
\textbf{Berkeley AI Research}%
}

\newcommand{\changes}[1]{{\color{blue}{#1}}~}

\newcommand{\methodName}[0]{ReLMM\xspace}

\newcommand{\refSection}[1]{Section \ref{#1}}
\newcommand{\refFigure}[1]{Figure~\ref{#1}}

\begin{document}

\maketitle

\begin{acronym}[ANOVA]
\acro{AGI}{artificial general intelligence}
\acro{ANOVA}[ANOVA]{Analysis of Variance\acroextra{, a set of
  statistical techniques to identify sources of variability between groups}}
\acro{ANN}{artificial neural network}
\acro{API}{application programming interface}
\acro{CACLA}{continuous actor critic learning automaton}
\acro{cGAN}{conditional generative adversarial network}
\acro{CMA}{covariance matrix adaptation}
\acro{COM}{centre of mass}
\acro{CTAN}{\acroextra{The }Common \TeX\ Archive Network}
\acro{DDPG}{deep deterministic policy gradient}
\acro{DeepLoco}{deep locomotion}
\acro{DOI}{Document Object Identifier\acroextra{ (see
    \url{http://doi.org})}}
\acro{DPG}{deterministic policy gradient}
\acro{DQN}{deep Q-network}
\acro{DRL}{deep reinforcement learning}
\acro{DYNA}{DYNA}
\acro{EOM}{Equations of motion}
\acro{EPG}{expected policy gradient}
\acro{FDR}{future discounted reward}
\acro{FSM}{finite state machine}
\acro{GAE}{generalized advantage estimation}
\acro{GAN}{generative adversarial network}
\acro{GPS}[GPS]{Graduate and Postdoctoral Studies}
\acro{HLC}{high-level controller}
\acro{HLP}{high-level policy}
\acro{HRL}{hierarchical reinforcement learning}
\acro{KLD}{Kullback-Leibler divergence}
\acro{LLC}{low-level controller}
\acro{LLP}{low-level policy}
\acro{MARL}{Multi-Agent Reinforcement Learning}
\acro{MBAE}{model-based action exploration}
\acro{MPC}{model predictive control}
\acro{MDP}{Markov Decision Processes}
\acro{MSE}{mean squared error}
\acro{MultiTasker}{controller that learns multiple tasks at the same time}
\acro{Parallel}{randomly initialize controllers and train in parallel}
\acro{PD}{proportional derivative}
\acro{PDF}{Portable Document Format}
\acro{PLAiD}{Progressive Learning and Integration via Distillation}
\acro{PPO}{proximal policy optimization}
\acro{PTD}{positive temporal difference}
\acro{RBF}{radial basis function}
\acro{ReLU}{rectified linear unit}
\acro{RCS}[RCS]{Revision control system\acroextra{, a software
    tool for tracking changes to a set of files}}
\acro{RL}{reinforcement learning}
\acro{SGD}{stochastic gradient descent}
\acro{Scratch}{randomly initialized controller}
\acro{SIMBICON}{SIMple BIped CONtroller}
\acro{SMBAE}{stochastic model-based action exploration}
\acro{SVG}{stochastic value gradients}
\acro{SVM}{support vector machine}
\acro{TL}{transfer learning}
\acro{TD}{temporal difference}
\acro{terrainRL}{terrain adaptive locomotion}
\acro{TLX}[TLX]{Task Load Index\acroextra{, an instrument for gauging
  the subjective mental workload experienced by a human in performing
  a task}}
\acro{TRPO}{trust region policy optimization}
\acro{UBC}{University of British Columbia}
\acro{UCB}{upper confidence bound}
\acro{UI}{user interface}
\acro{UML}{Unified Modelling Language\acroextra{, a visual language
    for modelling the structure of software artefacts}}
\acro{URDF}{unified robot description format}
\acro{URL}{Unique Resource Locator\acroextra{, used to describe a
    means for obtaining some resource on the world wide web}}
\acro{W3C}[W3C]{\acroextra{the }World Wide Web Consortium\acroextra{,
    the standards body for web technologies}}    
\acro{XML}{Extensible Markup Language}

\end{acronym}

\begin{abstract}
We study how robots can autonomously learn skills that require a combination of navigation and grasping. While reinforcement learning in principle provides for automated robotic skill learning, in practice reinforcement learning in the real world is challenging and often requires extensive instrumentation and supervision. Our aim is to devise a robotic reinforcement learning system for learning navigation and manipulation together, in an \textit{autonomous} way without human intervention, enabling continual learning under realistic assumptions. Our proposed system, ReLMM, can learn continuously on a real-world platform without any environment instrumentation, without human intervention, and without access to privileged information, such as maps, objects positions, or a global view of the environment. Our method employs a modularized policy with components for manipulation and navigation, where manipulation policy uncertainty drives exploration for the navigation controller, and the manipulation module provides rewards for navigation. We evaluate our method on a room cleanup task, where the robot must navigate to and pick up items scattered on the floor. After a grasp curriculum training phase, ReLMM can learn navigation and grasping together fully automatically in around 40 hours of autonomous real-world training.
\end{abstract}

\footnote[]{* denotes equal contribution}

\section{Introduction}
\begin{wrapfigure}{r}{0.35\textwidth} 
    \vspace{-0.5cm} %
    \centering
    \includegraphics[trim={0.0cm 0.0cm 0.0cm 0.0cm},clip,width=0.95\linewidth]{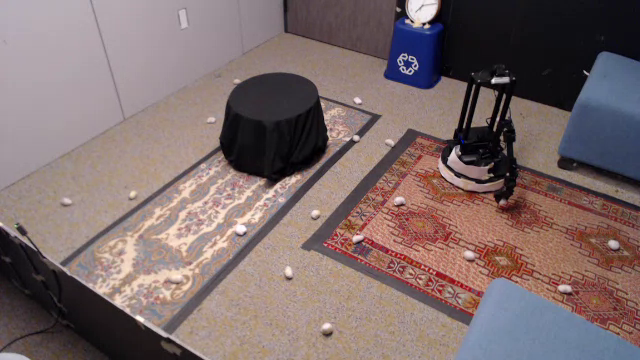}
    \caption{
    \methodName enables learning mobile manipulation skills autonomously in the real world, using only on-board sensing.}
    \label{fig:task-outline}
    \vspace{-0.5cm}
\end{wrapfigure}
Learning-based approaches have the potential to bring robots into open-world environments with end-to-end vision-based policies that can perform tasks without external instrumentation, AR-tagging, or expensive sensors. Such learning systems can be particularly beneficial for mobile manipulators, which perform tasks while navigating through open-world environments that are not practical to instrument or map out in advance. Unfortunately, in practice, most real-world reinforcement learning (RL) systems require careful environment instrumentation and human supervision or demonstrations during the training process to ensure that the robot performs effective exploration and resets between trials, making them difficult to apply to the kinds of open-world domains where we might want mobile manipulators to operate. Such systems might require specially installed infrastructure that provides explicit resets~\cite{chebotar2017path, Zhu2020, nagabandi2020deep, agrawal2016learning}, or the presence of a person providing resets and monitoring the learning process~\cite{chebotar2017combining, ghadirzadeh2017deep}, and cannot simply be dropped into a natural environment and continue learning.

To address this issue and make it possible for mobile manipulators to learn with RL directly in the real world, we propose a system for learning mobile manipulation skills without instrumentation, demonstrations, or manually-provided reset mechanisms. We aim to produce a system that enables a robot to learn autonomously in settings such as homes and offices, such that anyone could simply place the robot down, start the learning process, and return to a trained robot. This goal dictates several constraints that shape our method: (1) the robot must learn entirely from its own sensors, both to select actions and to compute rewards; (2) the entire learning process must be efficient enough for real-world training; (3) the robot must be able to continually gathering data at scale without human effort. A system that meets these requirements would not only be able to learn skills in open-world settings, but could also continue to improve throughout its lifetime: when learning can be performed practically without instrumentation, there is no reason to stop the learning process at deployment time, and the robot keep getting better and better at its given task perpetually. Our aim is not to propose the best possible system for solving any particular task. Rather, we aim to show how to create a real-world reinforcement learning system that enables learning mobile manipulation skills entirely from real-world interaction, with minimal human intervention.

\begin{figure*}
    \centering
            \includegraphics[trim={0.0cm 0.0cm 0.0cm 0.0cm},clip,width=0.155\linewidth]{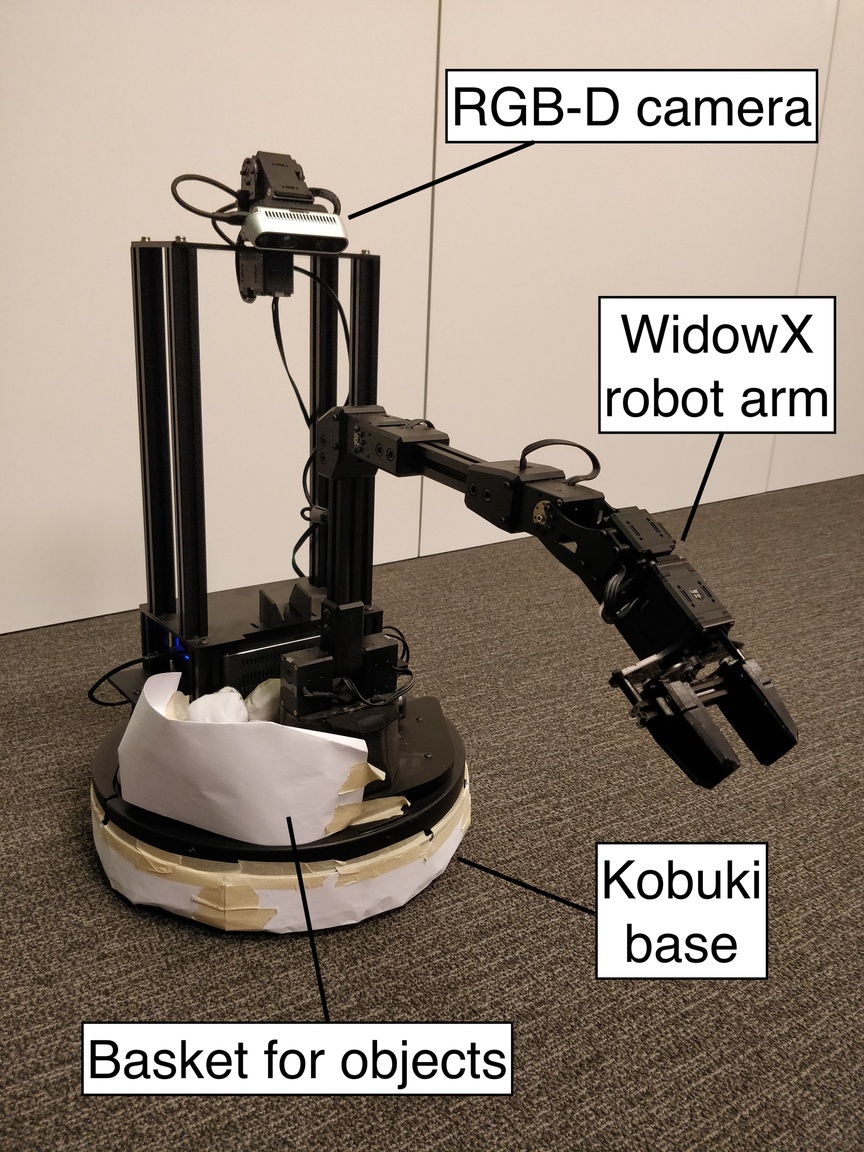}
    \includegraphics[width=0.75\textwidth]{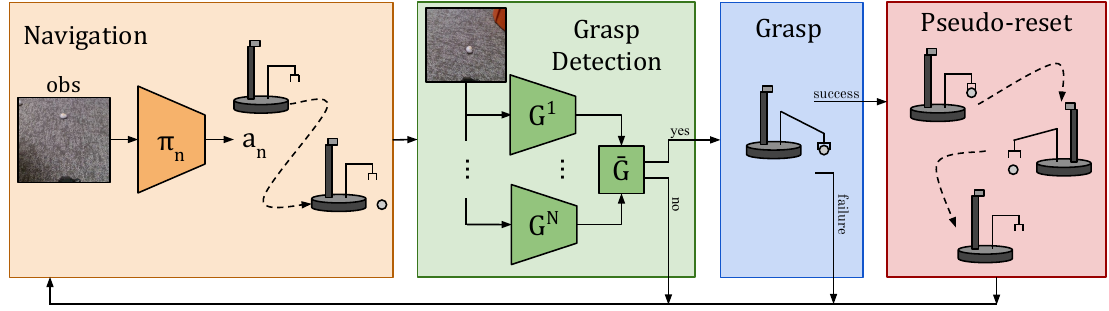}
    \caption{\textbf{Method overview.} ReLMM partitions the mobile manipulator into a navigation policy and grasping policy. Both policies are rewarded when an object is grasped. We use an ensemble of action-conditioned grasp success prediction functions estimate the success of a potential grasp better and use the uncertainty as an exploration bonus. If grasp success is likely, a grasp action is sampled from the grasp predictors and executed. If the grasp is successful during training, the robot executes a pseudo-reset by placing the collected object back down in a random location.}
    \label{fig:System}
    \vspace{-.7cm}
\end{figure*}

Our contribution is a system for autonomously training a mobile manipulation robot that satisfies the above constraints, which we call Reinforcement Learning for Mobile Manipulation (ReLMM). We apply ReLMM to the task of collecting items scattered across a room. Our system learns directly from on-board ego-centric camera observations and uses proprioceptive grasp sensing to assign itself rewards. To ensure the learning process is sample efficient and to facilitate exploration, we split the robot's controller into separate navigation and grasping neural network policy modules that choose when to act based on their predicted value and are continuously trained together for the same objective: successfully grasping objects in the environment. %
Separating the policies enables the use of uncertainty-based exploration for the grasping module, which uses an ensemble of Q-functions to explore grasp actions efficiently. %
Lastly, to reduce the need for humans to provide interventions in the form of resets, we develop an autonomous resetting behavior where the robot re-arranges the environment as it learns, so as to continually create new arrangements of objects for the agent continually ``practice.'' Together, these components plus a brief grasping curriculum enable a system that can operate in real-world environments, learning how to navigate and grasp from its own collected experience, and mastering room cleanup tasks with about 40-60 hours of autonomous interaction.
Videos are available at \url{https://sites.google.com/view/relmm}

\section{Related Work}
\label{sec:related_work}

Robotic mobile manipulation tasks pose a number of unique challenges~\cite{nishida2006development}. Many prior methods have addressed these challenges by requiring human effort for instrumentation and state estimation~\cite{stilman2005navigation,Stulp2009-jg,Stulp2009-kf,kaelbling2012unifying}, hand-coded controllers~\cite{Lehner2018}, or demonstrations~\cite{Stulp2009-kf,Walter2012-pd,Welschehold2017-zg}.\
Several methods also require external instrumentation such as top-down camera views, oracle knowledge of object pose, or precomputed navigation maps~\cite{DBLP:journals/corr/abs-1810-03400,wang2020mobile,Wu-b-RSS-20}. textcolor{Mobile manipulation has also seen benefits from combining learning and planning for more efficient exploration in simulation~\cite{kindle2020whole,mittal2021articulated,Honerkamp2021LearningKF}, which allow for safe and effective behavior in the real world if object or goal locations are provided. While these choices are pragmatic, they do not address the problem of learning continually in uninstrumented real-world settings and require a simulated world to train in.}
In contrast, our proposed system is aimed at enabling reinforcement learning directly on the real hardware that is maximally autonomous, and does not require external instrumentation.

Hierarchical reinforcement learning has been shown to learn interactive navigation and object rearrangement tasks in simulation operating from first-person view~\cite{li2020hrl4in,Xia2020-vg}, but require millions of timesteps to train and have not been demonstrated in the real world. Our system uses a similar hierarchical structure to these, but targets accessible real world learning by leveraging policy uncertainty estimates and curriculum learning for realistic sample efficiency.

While RL-based methods allow agents to improve via interaction, enabling robots that can learn outside of instrumented laboratory settings is difficult. Such settings lack episodic structure and well-shaped reward functions~\cite{sridharan2007structure,krajnik2016persistent, biswas2013cobot,rosenthal2011learning, pineau2003towards} that are important for success~\cite{lesort2020continual, coreyes2020ecological}. 
Large-scale uninterrupted deployment and training in the real world has been studied in several prior works~\cite{ thrun2001robust, fox1999mapping, kahn2017gcg, scherer2008ijrr, bruce2018kilometer, dronet2018, kahn2020badgr}.
While many of these works train robots in the real world without human interventions, they focus on navigation without manipulation.
Large-scale real-world learning has also been successful for robotic grasping in laboratory settings. By letting robots learn from their own experience, prior approaches have shown that robots can learn to grasp from images~\cite{lenz2015deep, pinto2016supersizing,levine2018learning, kalashnikov2018qt} or point clouds~\cite{ten2017grasp, mahler2017dexnet}. These methods show real world learning can lead to robust robotic behavior, but they do not tackle challenging mobile manipulation tasks. 

Gupta et al. show an approach to training a grasping policy on a mobile manipulator, but unlike our work, do not attempt to learn to navigate~\cite{gupta2018robot}. Wang et al. train a mobile manipulator with RL in simulation, but require known object pose, dense rewards, and episodic resets to a single initial state~\cite{wang2020mobile}. In contrast, our approach learns entirely from a sparse reward and onbaord sensors.
Past approaches to learning without episodic resets have focused on stationary manipulation or simulation~\cite{Eysenbach2018Leave, 7354297, Zhu2020}. In this work we explicitly focus on how \emph{mobile manipulation} robots can learn without external sources of resets or state estimation in the real world and lay out a set of design decisions that makes this process a practical one for acquiring robot skills.

\section{Preliminaries}

In this work, we use reinforcement learning (RL) as a general purpose algorithm for learning robotic behaviors. Reinforcement learning has the advantage of being able to operate on autonomously collected data, and enables the robot to improve through trial and error. To this end, the mobile manipulation task is formulated as a partially observed Markov decision processes (POMDP) with an observation space $\mathcal{O}$ of first person RGB images,
state space $\mathcal{S}$, action space $\mathcal{A}$, reward function $r(s_t, a_t)$, transition dynamics $\mathcal{P}(s_{t+1}|s_t, a_t)$, observation probability $\mathcal{P}(o_t|s_t)$, a discount factor $\gamma$, and an initial state distribution $\rho(s_0)$. The goal of reinforcement learning is to learn a control policy $\pi(a_t|o_t)$ that can determine which actions to take in each observation such that the expected sum of rewards is maximized. This objective can be written as 
$
    J(\pi) = \mathbb{E}_{a_t \sim \pi(o_t), s_t, o_t \sim \mathcal{P}}\left[\sum_t \gamma^t r(s_t, a_t)\right],
$ as for a standard RL problem.

\section{ReLMM: RL For Mobile Manipulation}
\label{sec:framework}

\begin{wrapfigure}{r}{0.5\textwidth}
\vspace{-0.45cm}
\begin{minipage}{.5\textwidth}
\begin{algorithm}[H]
\caption{TrainGrasp($G^{1}, ..., G^{M}$, $\mathcal{D}_g$, $N$, $st$)}
\label{algo:slap_grasp}
\begin{algorithmic}[1]
\For{t = 0, \dots, $N$ steps} 
    \State Get grasp observation $\Tilde{o}$.
    \State Sample $a_g \sim \pi_g(\cdot|\Tilde{o})$. // see Equation~\ref{eq:grasping-policy}
    \State Perform grasp $a_g$, receiving $r_g = 0$ or $1$.
    \State Store $(\Tilde{o}, a_g, r_g)$ in $\mathcal{D}_g$.
    \State {Update $G^{1}, ..., G^{M}$ on $\mathcal{D}_g$}
    \State{\textbf{if} $st =$True, Drop object if held.}
    \State{\textbf{elseif} $r_g=1$ and $st =$False, \textbf{return} $r_g$}
\EndFor %
\State{\textbf{return} 0}
\end{algorithmic}
\end{algorithm}
\end{minipage}
\vspace{-0.2cm}
\end{wrapfigure}
We develop the \methodName system to enable training mobile manipulation robots with RL. While we specifically apply it to a room cleaning task, in principle \methodName could be used to learn other mobile manipulation tasks as well. Each component of \methodName is chosen in order to maximise the autonomy of learning while retaining the sample efficiency needed to train in the real world. %
The specific task that we study in our experiments involves training a robot to quickly navigate around in a room with obstacles, physically pick up many objects, and place them in a basket mounted on the robot, as shown in ~\autoref{fig:task-outline} and \ref{fig:real-robot-results}.

Our system provides for efficient autonomous learning by decomposing the policy into grasping and navigation policies, using an ensemble of grasping models to explore based on uncertainty, automatically rearranging the environment after successful grasps, and using a curriculum to bootstrap and stabilize the concurrent training of both policies.
Our final system can learn room cleaning skills in a number of different room configurations in $\sim40$ hours directly in the real world.

\subsection{Grasping Policy Training}
\label{par:grasping}

As noted previously, we decomposed the control problem into grasping and navigation. This gives the manipulation policy two objectives: given an image observation, accurately model the robots chance of success should it attempt a grasp and,
if so, select an appropriate action to maximize success. We obtain the former by training an ensemble of grasping policies and using their uncertainty to efficiently explore grasping. For choosing \emph{how} to grasp, the policy must learn with sufficiently low sample complexity so as to make real-world training feasible. To reduce the complexity of the exploration problem we formulate grasping as a discrete single-step top-down action. The grasp policy $\pi_g(a_g|o_g)$ is parameterized with the action $a_g$ discretized in the x-y plane, and the observation $o_g$ corresponding to a RGB image from the robot's camera. Such single-step action selection formulations are amenable to more efficient training than more complex multi-step tasks~\cite{NEURIPS2018_d3b1fb02,Wu-RSS-20}.

Framed in this way, the grasping task corresponds to a contextual multi-armed bandit problem. Specifically, we train grasping policies that, given an image, predict the likelihood of grasp success for each action.%
We use a soft-max over the action values to sample actions in proportion to their exponentiated probability of success. %
To create the ensemble, we train $M=6$ independent grasp policies, $G^{1} \ldots  G^{M}$ that are each by minimizing the cross-entropy loss on the same dataset:
\begin{equation}
\label{eq:grasping-loss}
    \mathcal{L}_g^{i} = \mathbb{E}_{(o_g, a_g, r_g)\sim \mathcal{D}_g}[-r_g\log G^{i}(o_g, a_g) - 
    (1-r_g)\log (1 - G^{i}(o_g, a_g))].
\end{equation}
Here, $r_g$ is 1 when the robot successfully grasps an object, which is determined by presenting the gripper to the onboard camera.
The grasping exploration policy is formed by constructing a Boltzmann distribution from optimistic estimates of grasp success, where the mean estimate from the ensemble, $\mathbb{E}[G^{i}(o_g, a_g)]$, is modified by adding a multiple of the ensemble variance $\sigma(G^{i}(o_g, a_g))$, which we expect to be larger for actions where success is more uncertain:
\begin{equation}
\label{eq:grasping-policy}
    \Tilde{G}(o_g, a_g) = \alpha\mathbb{E}[G^{i}(o_g, a_g)] + \beta\sigma(G^{i}(o_g, a_g)),
\end{equation}
with 
$
    \pi_g(a_g|o_g) \propto \exp(\Tilde{G}(o_g, a_g)).
$
The expectation and standard deviation are taken over the ensemble, and $\alpha,\beta\geq 0$ are hyperparameters. Other equivalent multi-step off-policy or contextual bandit algorithms can be used to train the grasping policy. However, we show in~\refSection{sec:simulation_analysis} that they are not as sample efficient for the mobile manipulation task in this work. 
Algorithm~\ref{algo:slap_grasp} describes the grasp training process in further detail.

\subsection{Navigation Policy Training}
\label{par:nav}
For navigation, the policy must be able to control the mobile base to approach objects in a way that the current grasping policy can succeed. The navigation policy $\pi_n(a_n|o)$ outputs the action $a_n$ that controls the forward and turn velocities of the mobile robot base.

At every time step, the agent has to decide whether to perform a grasp or not by balancing the opportunity of receiving reward, the chance to collect novel data for the grasping policies, and the cost of wasting a timestep if there is no object within reach. %
This balancing act is done by reusing the same uncertainty measure described in Equation \ref{eq:grasping-policy} 
 when choosing whether to attempt a grasp. The probability of whether to attempt a grasp is
\begin{equation}
\label{eq:do-grasp-prob}
    \mathbb{P}[\text{grasp}|o] = \max_{a_g} \Tilde{G}(o, a_g).
\end{equation}
Under this design, the navigation policy $\pi_n(a_n|o)$ continues to experience observations and output navigation actions, and at every step the choice of whether to perform a grasp or not is made by sampling from grasp success probability of the current grasp model  $\mathbb{P}[\text{grasp}|o]$
defined in Eqn~\ref{eq:do-grasp-prob}. When the model decides attempting a grasp is worth the risk, the robot executes a grasp and evaluates the outcome to provide the navigation agent a reward. From the perspective of the navigation policy, the choice of whether to grasp or not is a part of the inherent dynamics of the environment. %
We compare two possible rewards $r_n$ for the the navigation policy. The first option is directly optimizing for the task by rewarding the navigation when a grasp is successful (i.e. where the current grasp ensemble has high performance): $r_n(o) = r_g(o) - 1$. The second option is to reward the navigation for reaching states that the current grasp ensemble will choose to grasp at, which is a function of its mean and uncertainty: $r_n(o) = (\mathbb{P}[\text{grasp}|o]-1)$ that we use to \textit{relabel} states without successful grasps during SAC policy update. As this reward function only depends on the grasp ensemble and not on actual grasp success, this reward is computed at every policy update step and in~\autoref{fig:abblation-results} we show how it can improve sample efficiency.
The RL objective for navigation is 
$
    \max_{\pi_n} \mathbb{E}_{\pi_n(a_n|o_t)}\left[\sum_{k=0}^{\infty} \gamma^{k} r_n(o_{t+k})\right.]
$
We train the navigation policy for this objective using soft actor critic (SAC)~\cite{haarnoja2018soft,haarnoja2019learning}.

\subsection{Training with Autonomous Pseudo-Resets}
\label{par:reset}
\begin{wrapfigure}{r}{0.52\textwidth}
\vspace{-0.5cm}
\begin{minipage}{.52\textwidth}
\begin{algorithm}[H]
\caption{\methodname (with Stationary Curriculum)}
\label{algo:slap}
\begin{algorithmic}[1]
\label{alg:slap-grasp-train}
\State{Hyperparameters: $M$, $N_\text{st}$, $N_\text{grasp}$, \textit{relabel}}
\State{Init: function estimators $\pi_n$, $G^{1}, \ldots, G^{M}$}.
\State{Replay buffers $\mathcal{D}_n = \{\}$, $\mathcal{D}_g = \{\}$}
\State{TrainGrasp($G^{1}, .., G^{M}$, $D_g$, $N_\text{st}$, True)}
\For{$t = $ 0, \dots, $T$ steps}
    \State Get navigation observation $o_t$
    \State Sample $a_n \sim \pi_n(\cdot|o_t)$ and perform $a_n$
    \If{ \textit{uniform}() $\leq$ $\mathbb{P}[\text{grasp}|o_t]$}
        \State {$r_{g}\!= $TrainGrasp($G^{1},\! .., G^{M}$,$D_g$,$N_\text{grasp}$,False)}
    \Else { $r_g = 0$}
    \EndIf
    \If{\textit{relabel}}
        \State{Navigation reward $r_n = \mathbb{P}[\text{grasp}|o_t] - 1$}
    \Else{ $r_n = r_g - 1$}
    \EndIf
    \State Get next navigation observation $o_{t+1}$
    \State Store $(o_t, a_n, r_n, o_{t+1})$ in $\mathcal{D}_n$.
    \State Update $\pi_n$ with $\mathcal{D}_n$ using SAC.
    \State Pseudo-reset
\EndFor
\State {\small{\textbf{end for}}}
\end{algorithmic}
\end{algorithm}
\end{minipage}
\vspace{-0.5cm}
\end{wrapfigure}
    While the training schemes described above allow \methodName to learn efficiently, both the contextual bandit grasping formulation and the navigation training setup requires an episodic training setup, where the environment is reset between trials, for example by replacing the objects into the world at random positions. To enable the robot to learn mobile manipulation skills without manually provided resets, we construct an automated pseudo-reset system that allows our method to learn autonomously without human intervention. 
After a successful grasp, the environment would ideally be reset by relocating the object to a new, randomly selected location. In stationary bin grasping setups, this can be automated simply by dropping the object back into the bin. However, for a mobile manipulator, dropping the object back to where it was grasped would promote policies that fail to search for new objects, and simply remain in the same location. To force the policies to learn to grasp in varied situations, we perform an automated random pseudo-reset, by commanding random navigation actions while the robot is holding the object, placing down the object in this new location, and then navigating randomly away. This ensures the robot will not always be near an object during training and must instead learn to seek objects out. We outline the overall algorithm used for training in Algorithm~\ref{algo:slap}.

\subsection{Training Curricula}
The separate grasping and navigation policies lend themselves naturally to curricular training, where the grasping policy, which needs to have some successes to provide rewards to the navigation policy, can be prioritized at the beginning of the learning process. We propose two types of curricula, which both break the potential ``chicken-and-egg" training problem of providing poor reward signals for navigation from an untrained grasping model.

\noindent \textbf{Stationary curriculum.} The simplest curriculum is to place a single object in front of the robot and run Algorithm \ref{algo:slap_grasp} with $st=\text{True}$ for $N_\text{st}=2000$ steps. After each successful grasp, the robot places the object down randomly. If the object is knocked into an area the robot can not reach (because the base is stationary), which happens about $5\%$ of the time, a human observer must push the object back into the graspable area. This curriculum is very time efficient, but does require occasional human intervention.

\noindent \textbf{Autonomous curriculum.} For fully autonomous learning, we develop a training curriculum that favors collecting grasping data early on by performing a high number of grasps at the beginning of training, according to hyperparameters $N_{start}$,  $N_{stop}$, and $N_{max}$. The robot attempts $N_{grasp} = N_{start}$ grasps after every navigation step. If the robot succeeds at grasping an object it will practice with this object until $N_{grasp} = N_{stop}$ unsuccessful grasps. 
This initial automatic grasping curricula ends when a total of $N_{max}$ grasps are complete. More details on the grasping curriculum algorithm are in Appendix~\ref{sec:grasp-curriculum}.

\section{Robotic Mobile Manipulator: System Overview}
\label{sec:system-overview}

Our choice of robotic system reflects the need for a robust robotic platform that can operate autonomously for long periods of time, and is unlikely to cause damage to itself and its surroundings. Ensuring safety during autonomous operation is itself a significant research challenge, which is outside of the scope of this work. Therefore, we utilize a small-scale low-cost mobile manipulation platform based on the LocoBot design~\cite{gupta2018robot,murali2019pyrobot}, shown in Figure~\ref{fig:System} (left), which consists of an iClebo Kobuki mobile base and a WidowX200 5-DoF arm. The robot sensors include an Intel RealSense D435 camera at the top of the robot, as well as bump sensors on the base. We use an onboard Intel NUC to command the robot, and connect wirelessly to a server for data processing and training. Random exploration is generally safe with the LoCoBot as it is small, light, and will automatically stop the arm's motors if they encounter resistance. We also use the depth camera output to automatically avoid collisions, as described in \autoref{app:collision}.
The robot learns in a real-world office space, with varied lighting conditions, distractors, and surface textures. Our experiments utilize small objects that the robot can feasibly pick up, which consist of socks and toys -- a sampling of objects one may want a robot to pick up off the floor.

To control the robot, we separately command the iClebo mobile base and the WidowX200 robotic arm with the corresponding navigation and a grasping policies. The navigation policy stops the robot during a grasp. For grasping, we use a directed end-effector control space. Assuming the floor is flat, the vertical position for grasping is always chosen to be just above the floor, and the learning algorithm chooses a point in $X-Y$ space in front of the robot to perform a grasp at. This chosen position then dictates where to move the gripper using inverse kinematics. 
Details on the learned network architectures are in Appendix~\ref{sec:model-architecture}.

\section{Experimental Results}
\label{sec:results}

Our experiments aim to evaluate our autonomous reinforcement learning system in a number of real-world environments, as well as to provide ablation experiments and analysis in simulation. In particular, we aim to study the following questions: (1) Can \methodName learn autonomously in the real world? (2) How does the control hierarchy affect learning performance? (3) How does \methodName compare to other policy designs and prior methods? Our goal is to study real-world reinforcement learning, rather than necessarily provide the best possible solution to the room clearing task, and hence we design our system to be general, with only the reward determining the task.
\begin{figure*}[t]
    \centering
    \includegraphics[trim={0.0cm 0.0cm 0.0cm 0.0cm},clip,height=0.17\linewidth]{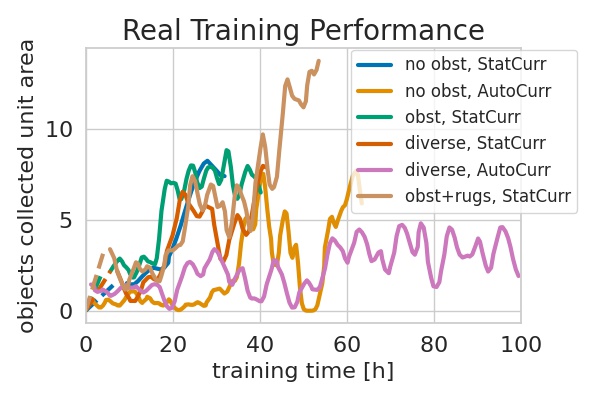}
    \includegraphics[trim={15.0cm 0.0cm 3.0cm 0.0cm},clip,height=0.17\textwidth]{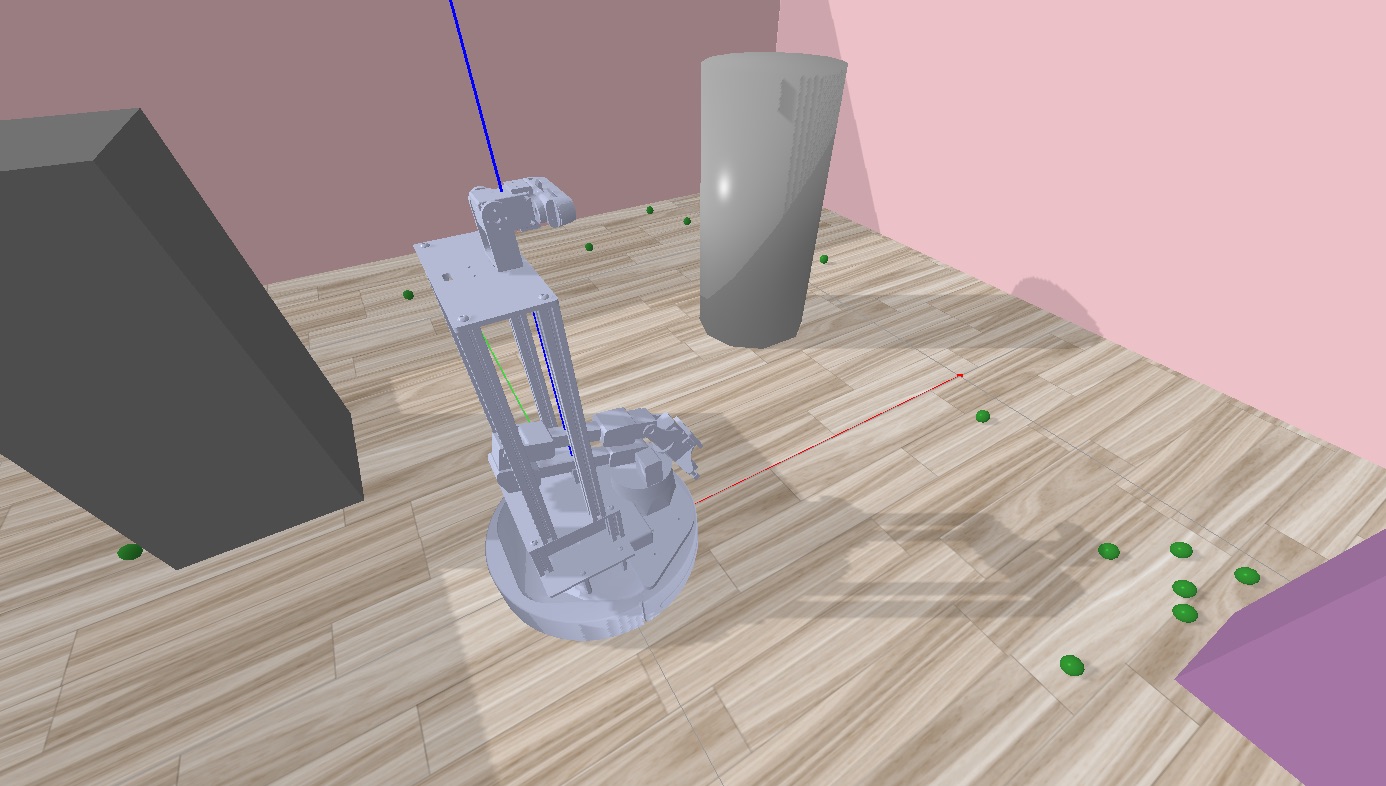}
    \includegraphics[trim={0.0cm 0.0cm 0.0cm 0.0cm},clip,height=0.17\linewidth]{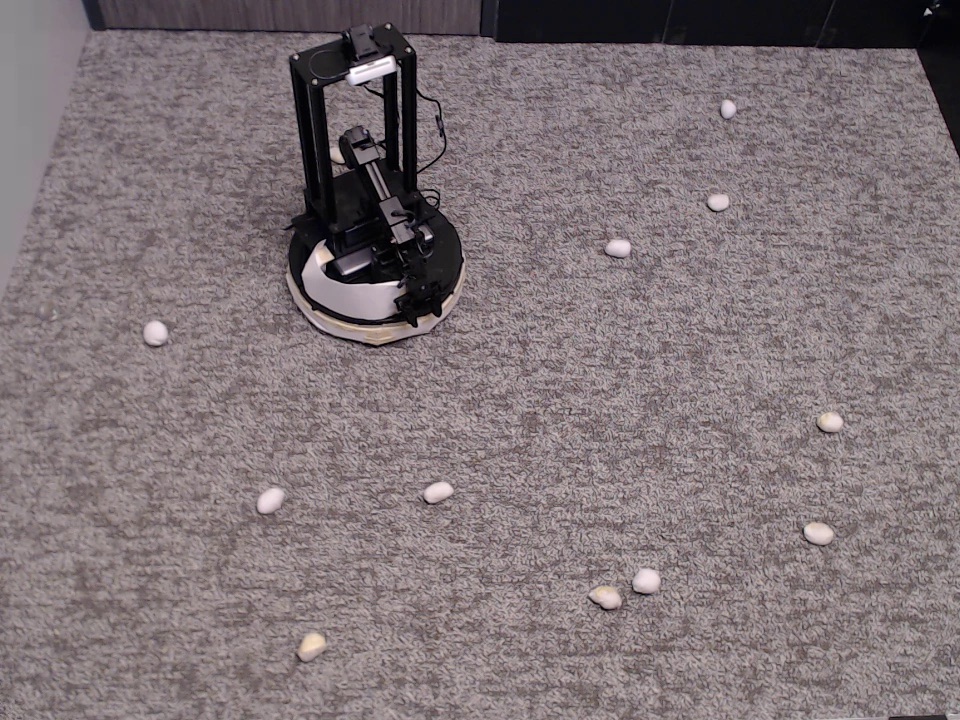}
    \includegraphics[trim={0.0cm 0.0cm 0.0cm 0.0cm},clip,height=0.17\linewidth]{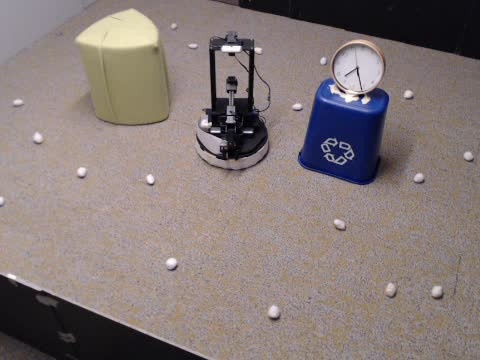} \\
    \includegraphics[trim={0.0cm 0.0cm 0.0cm 0.0cm},clip,height=0.17\linewidth]{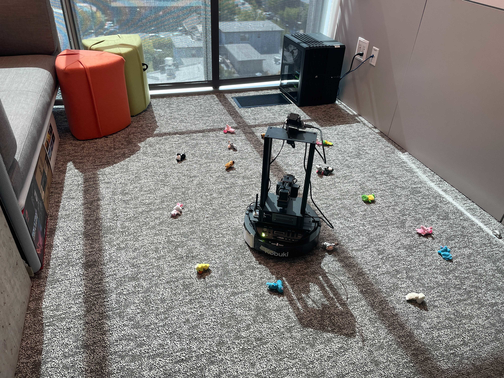}
    \includegraphics[trim={5.0cm 0.0cm 1.0cm 0.0cm},clip,height=0.17\linewidth]{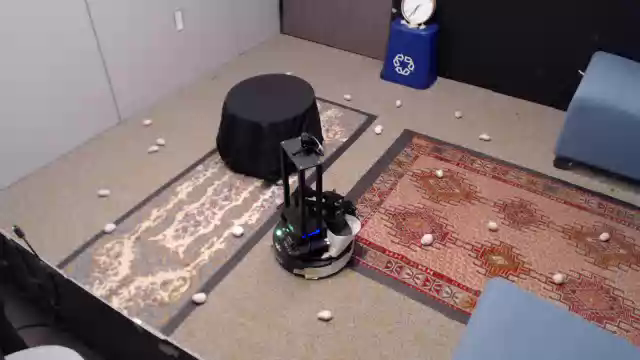}
    \includegraphics[trim={5.0cm 0.0cm 1.0cm 0.0cm},clip,height=0.17\linewidth]{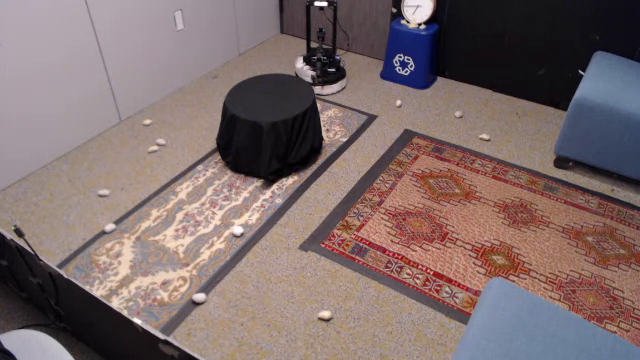}
    \includegraphics[trim={5.0cm 0.0cm 1.0cm 0.0cm},clip,height=0.17\linewidth]{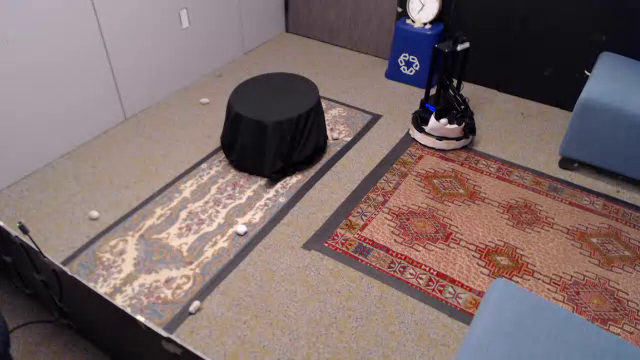}
    
    \caption{{\footnotesize \textbf{Top Left}: Learning curves for training the real robot showing the number of objects grasping in the last $15$ min of operation, which is $\sim250$ actions. This metric are divided by the rooms' surface areas to make them comparable. \textbf{Top mid-left}: The simulated environment, which we run with and without obstacles. \textbf{Right \& Bottom}: Snapshots from the evaluation tasks without obstacles in a $4m^2$ room (top-mid-right, \textit{no obst}, bottom-left, \textit{diverse}), with obstacles \textit{obst} in a $9m^2$ room (top-right), and with obstacles and rugs \textit{obst+rugs} in a $10m^2$ room (bottom-right). Videos are available at  \url{https://sites.google.com/view/relmm/home}
    }
    }
    \label{fig:real-robot-results}
    \vspace{-0.5cm}
\end{figure*}
\noindent \textbf{Experiment details}
For all real-world experiments, the entire training procedure is performed in the real world on the LoCoBot platform, with about $25$ to $50$ hours of training depending on the environment.
The training is autonomous, with the exception of the stationary curriculum (if used), and battery changes every $\sim5$ hours, during which time we may replace objects that become stuck near corners, or walls.

As our task is non-episodic, the policies are evaluated at the end of training. This is done by scattering the objects, executing the policy (without pseudo-resets) for 15 minutes, and counting the number of objects collected in that time. We use four real environments in our evaluation (no obstacles, with obstacles, with diverse objects, and with obstacles and rugs and a simulated environments, as shown in Figure~\ref{fig:real-robot-results}. Each room has a different size and number of objects, so we report the percent of objects collected in each 15 minute block of training, as shown in \autoref{fig:real-robot-results}.
In simulation, we plot the evaluation performance using 250 timesteps instead of 15 minutes.

\begin{wraptable}{r}{0.55\textwidth}
    \vspace{-0.5cm}
    \centering
    \setlength\tabcolsep{1.5pt} %
    {\footnotesize
    \begin{tabular}{|c|c|c|c|c|} \hline
       Env & no obst & obst & diverse & obst+rugs \\ \hline
        \textit{\methodName-StatCurr}  & $\mathbf{88 \pm 2}$ & $\mathbf{93 \pm 2}$ & $63 \pm 8$ & $\mathbf{78 \pm 6}$  \\ \hline 
        \textit{\methodName-AutoCurr} & $77 \pm 8$ & -- & $\mathbf{75 \pm 4}$ & -- \\ \hline
       \textit{Scripted}   & $75 \pm 4$ & $88 \pm 6$ & $56 \pm 3$ & $65 \pm 9$ \\ \hline 
      \textit{Rand nav}~\citep{gupta2018robot}  & $52 \pm 12$ & $38 \pm 10$ & $22 \pm 9$ & $20 \pm 7$ \\ \hline
       \textit{Rand all}  & $12 \pm 6$ & $2 \pm 2$ & $2 \pm 3$ & $5 \pm 4$ \\ \hline 
    \end{tabular}
    }
    \caption{{\footnotesize \small Percentage of objects that the robot collects during eval in each environment (shown in Figure~\ref{fig:real-robot-results-frames}) (higher is better). Each method is trained once per env, and evaluated 3 times. The numbers are mean and stddev of the 3 evaluations. \methodName outperforms the baselines by learning both grasping and navigation jointly, without requiring environment instrumentation. Due to \methodName-AutoCurr's slower learning, we only evaluate it in \textit{no obst} with \textit{diverse}. In \autoref{tab:real_hours} we provide the training time in each env.
    }
    }
    \label{tab:real_results}
    \vspace{-0.25cm}
\end{wraptable}
We compare our approach to several prior methods and baselines.
The baselines include: \textbf{Rand all}, where the navigation policy generates actions from $\pi_n \sim \mathcal{U}[-1, 1]^2$, and similarly the grasping policy chooses uniformly random from the space of discrete actions, as a lower baseline; \textbf{Rand nav}, a method similar to \citet{gupta2018robot} where the navigation policy is $\pi_n \sim \mathcal{U}[-1, 1]^2$, but the grasping policy is loaded from a run of stationary curriculum right after the initial stationary phase (i.e. with stationary pretraining only). These baselines disentangle the benefits we get from learning both the grasping and the navigation. We also include a hand-coded controller (\textbf{Scripted}) specifically engineered for this task that locates objects by thresholding the image pixels and grasps at their centroids, which provides a strong hand-designed baseline. For more details on the implementation of the scripted controller, see Appendix~\ref{sec:scripted-policy}.

\begin{wrapfigure}{r}{0.45\textwidth} 
\vspace{-0.5cm}
    \centering
    \includegraphics[width=0.85\linewidth]{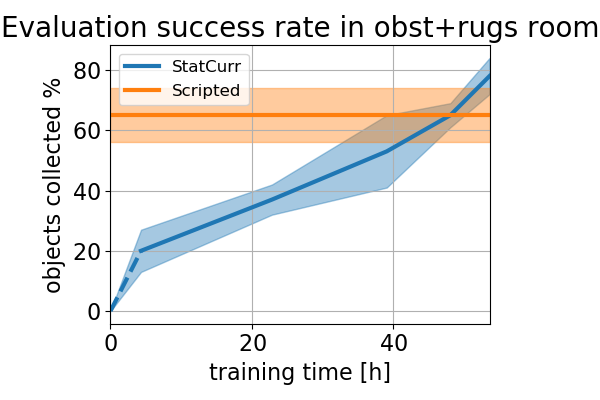}
    \vspace{-0.1cm}
    \caption{ \small
    Different checkpoints of \methodName in the obst+rugs room. Performance improves steadily with more training, and indeed is still increasing even at our final checkpoint, suggesting that lifelong training enables continual improvement for \methodName.
    \label{fig:eval_improvement}
    }
\vspace{-0.5cm}
\end{wrapfigure}

\begin{wraptable}{r}{0.475\textwidth}
    \vspace{-0.4cm}
    \centering
    \setlength\tabcolsep{1.5pt} %
    {\footnotesize
    \begin{tabular}{|c|c|c|c|c|} \hline
      Env & no obst & obst & diverse & obst+rugs \\ \hline
        \textit{\methodName-StatCurr}  & $25.1$ & $ 36.2$ & $35.7$ & $ 50.9$  \\ \hline 
        \textit{\methodName-AutoCurr} & $63.5$ & -- & $99.5$ & -- \\ \hline
    \end{tabular}
    }
    \caption{{\footnotesize \small Number of hours \methodName was trained to achieve the real robot performance noted in \autoref{tab:real_results}. 
    }
    }
    \label{tab:real_hours}
    \vspace{-0.5cm}
\end{wraptable}
\noindent \textbf{Real Robot Evaluation} 
Our real-world experiments evaluate how well \methodName can learn a mobile grasping policy autonomously in a variety of rooms.
We conducted separate experiments for each of the rooms shown in~\autoref{fig:real-robot-results}, which differ in terms of size, furniture, layout, and objects. %
\methodName can train using both the \textit{stationary} and \textit{autonomous} grasping curricula, with the \textit{autonomous curriculum} requiring less human effort, at the cost of increased training time. %
The result are summarized in \autoref{tab:real_results} with the training time given in \autoref{tab:real_hours}.
Examples of the learned behaviors are also shown in~\autoref{fig:real-robot-results-frames}, and are illustrated in more detail in the supplementary video. Due to the real world training constraints, we cannot run the evaluation protocol regularly during real-world training. However, we did run evaluation for several checkpoints for the ReLMM-StatCurr agent in the obst+rugs room, as shown in
\autoref{fig:eval_improvement}. The total number of hours trained in each environment is given in \autoref{tab:real_hours}. Frames from the evaluations are shown in \autoref{fig:real-robot-results-frames}.

\begin{figure*}[t]
    \centering
    \includegraphics[trim={0.0cm 0.0cm 0.0cm 0.0cm},clip,width=0.31\linewidth]{images/img001.jpg}
    \includegraphics[trim={0.0cm 0.0cm 0.0cm 0.0cm},clip,width=0.31\linewidth]{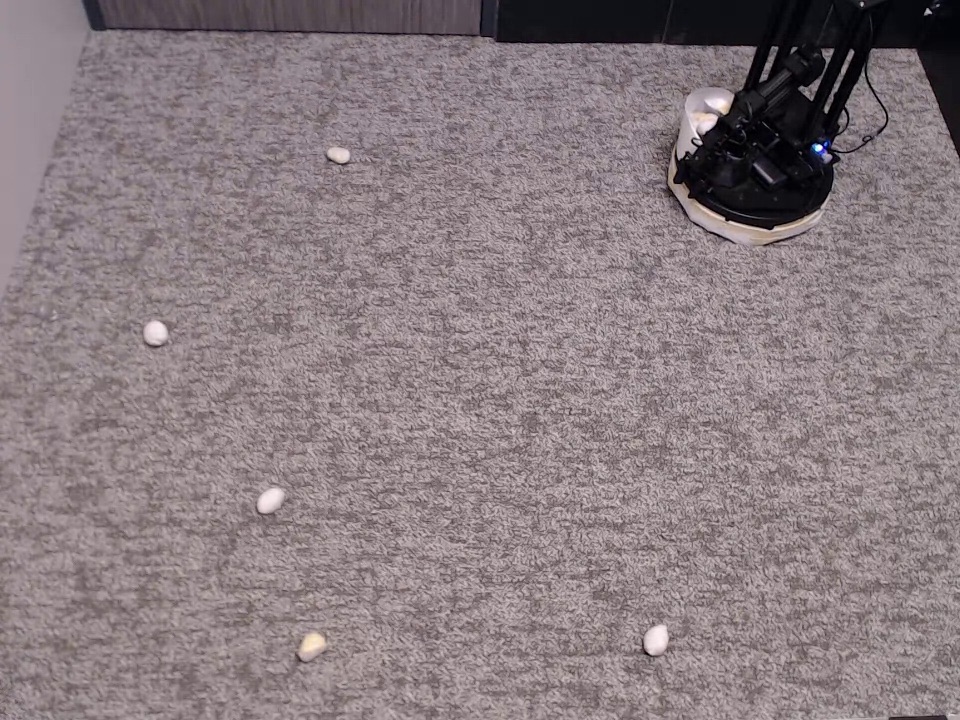}
    \includegraphics[trim={0.0cm 0.0cm 0.0cm 0.0cm},clip,width=0.31\linewidth]{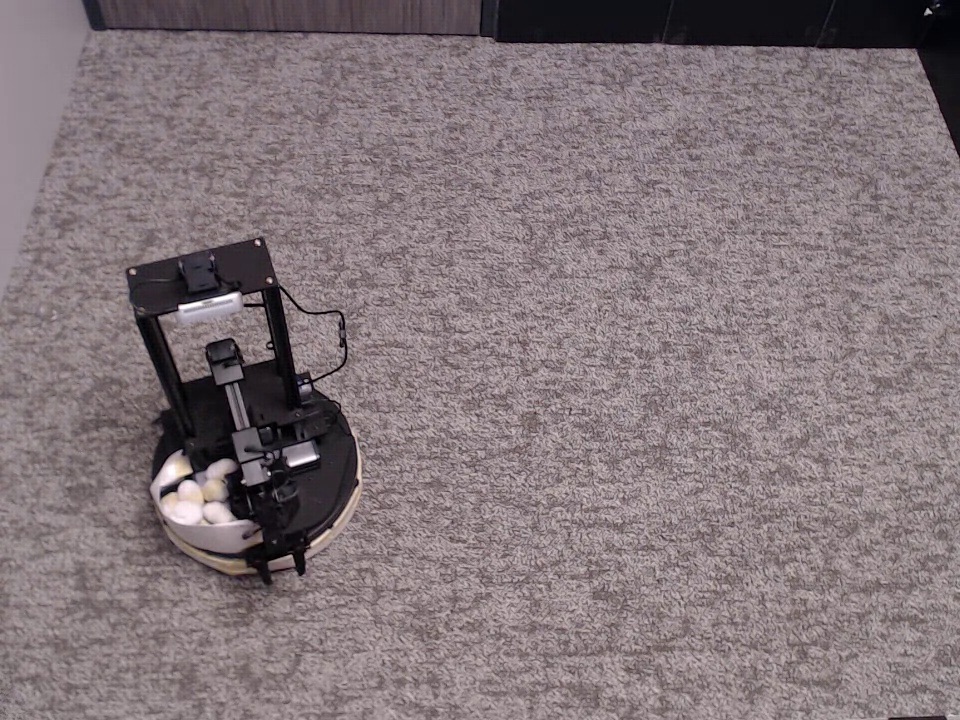} \\
    \includegraphics[trim={0.0cm 0.0cm 0.0cm 0.0cm},clip,width=0.31\linewidth]{images/obst_eval_f1.jpg}
    \includegraphics[trim={0.0cm 0.0cm 0.0cm 0.0cm},clip,width=0.31\linewidth]{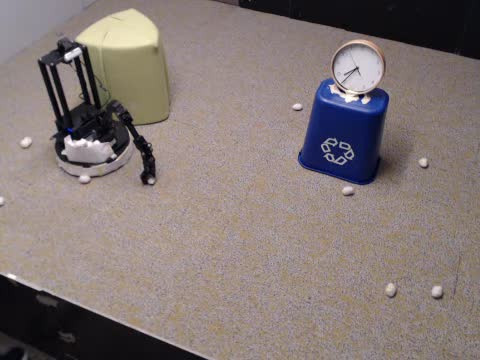}
    \includegraphics[trim={0.0cm 0.0cm 0.0cm 0.0cm},clip,width=0.31\linewidth]{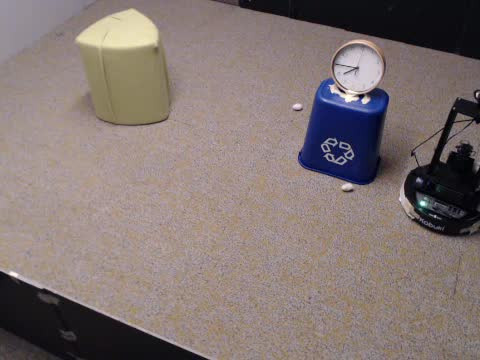} \\
     \includegraphics[trim={0.0cm 0.0cm 0.0cm 0.0cm},clip,width=0.31\linewidth]{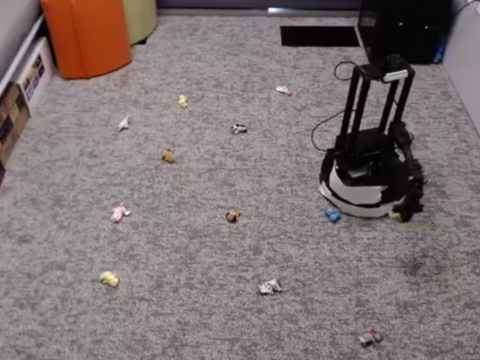}
    \includegraphics[trim={0.0cm 0.0cm 0.0cm 0.0cm},clip,width=0.31\linewidth]{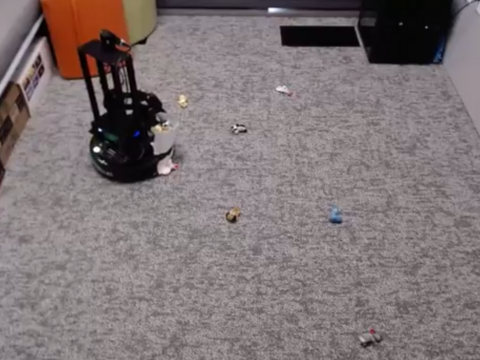} 
    \includegraphics[trim={0.0cm 0.0cm 0.0cm 0.0cm},clip,width=0.31\linewidth]{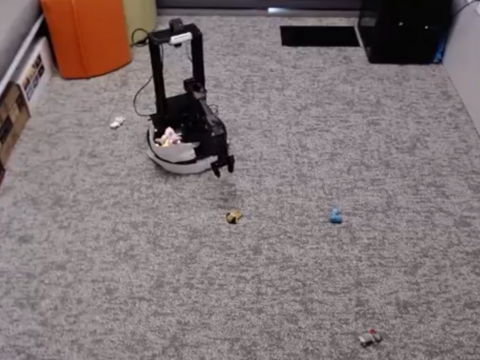} \\
    \caption{Frames from videos of the real robot evaluation in each room. The rooms starting from top down are \textit{no obst}, \textit{obst}, \textit{diverse}.
    }
    \label{fig:real-robot-results-frames}
    \vspace{-0.3cm}
\end{figure*}

\noindent To provide context for these results, we compare \methodName to the previously described baselines. 
The \textit{Rand all} and \textit{Rand nav}~\citep{gupta2018robot} baselines perform poorly, since neither method learns a directed navigation behavior, making it difficult to effectively navigate the room to collect objects, though \textit{Rand nav}~\citep{gupta2018robot} is effective at picking up the objects if it stumbles upon them. 

This illustrates the importance of an intelligent navigational strategy in these settings. The \textit{Scripted} baseline, which was specifically engineered for this task, performs reasonably in most settings, but still falls short of the policy learned by \methodName in all of the environments, and crucially cannot improve from these results autonomously.
This is particularly important in the \textit{diverse} and \textit{obst+rugs} environments where it is difficult to find a pixel threshold that works for all objects and all backgrounds, \methodName can automatically learn how to identify objects to grasp via interaction. In \autoref{fig:eval_improvement}, we plot the evaluation performance of our ReLMM-StatCurr agent learning in \textit{obst+rugs} for a few checkpoints, showing that our method is improving from its experience and is still improving after $50$ hours of training. This is important in open-world settings, where our approach would enable mobile manipulators to improve perpetually over the lifetime of their deployment.

\label{sec:method-comparison-details}

\begin{figure}[tb]
    \centering
    \includegraphics[trim={0.0cm 0.0cm 0.0cm 0.0cm},clip,height=0.2\linewidth]{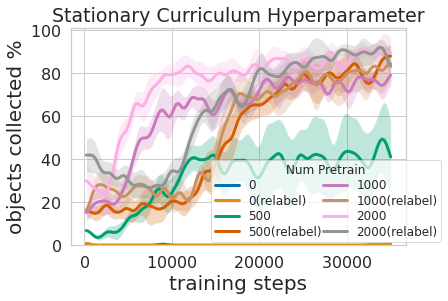}
    \includegraphics[trim={0.0cm 0.0cm 0.0cm 0.0cm},clip,height=0.2\linewidth]{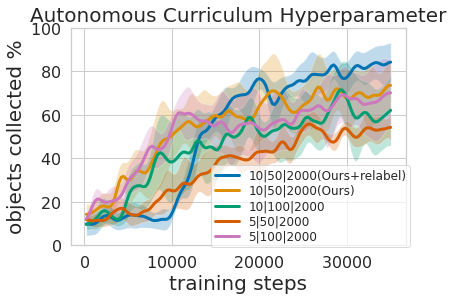}
    \includegraphics[trim={0.0cm 0.0cm 0.0cm 0.0cm},clip,height=0.2\linewidth]{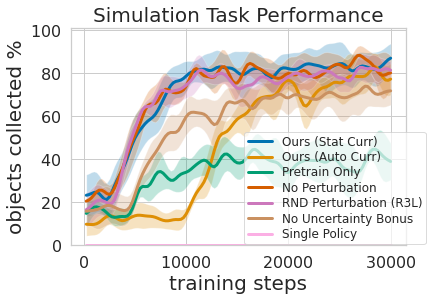}
    
    \caption{{\footnotesize \textbf{Performance in the simulated environment without obstacles.} Each plot is the mean and stddev over 3 random seeds. \textbf{Left}: Ablation of different number of grasp "pretraining" samples for the stationary curriculum ($N_\text{st}$) and navigation reward relabeling (described in \autoref{par:nav}) in simulation. Increased pretraining of the grasping policy improves overall mobile manipulation performance.
    \textbf{Center}: Analysis to find the best parameters for the \textit{autonomous} curriculum. With the proper settings ($N_{start}|N_{stop}|N_{max}$) the \textit{autonomous} curriculum can be almost as efficient as the \textit{stationary}, without requiring as much manual effort.
    \textbf{Right}: Ablation of \methodName, all use the stationary curriculum and no relabeling except for \textit{Ours (AutoCurr)}. We find that the uncertainty bonus and joint training are critical components of our system.%
    }
    }
    \label{fig:abblation-results}
    \vspace{-0.25cm}
\end{figure}

\noindent \textbf{Simulation Analysis}
\label{sec:simulation_analysis}
To study questions (2) and (3), we perform a detailed ablation analysis in simulation.
As discussed in~\autoref{app:results}, we find that a discretized grasping policy learns with $\frac{1}{3}$ the samples compared to a continuous one. 
In \autoref{fig:abblation-results}, we show the performance of various ablations of our system in the simulated environment with the \textit{stationary curriculum}.
First, we ablate the policy decomposition by training a \textit{single policy} with reinforcement learning using grasp success as the reward. The joint action space poses a much larger exploration problem, and the policy is unable to make headway on the task in a reasonable number of samples. This indicates that the hierarchical decomposition used by \methodName significantly improves training performance.
Next we see that freezing the grasp policy after the stationary phase (\textit{Pretrain Online}) and only training the navigation is much worse than training both together. This illustrates the interplay between the two policies, as the navigation is dependent on the grasping for obtaining rewards.
As shown in the plot, training the grasp policy without the uncertainty bonus (i.e. $\beta=0$) leads to significantly poorer performance for \methodName, as it is less incentivized to explore. Finally, we see that the autonomous curriculum with reward relabeling can get similar final performance as the stationary curriculum with less human intervention at the cost of longer training time.

Lastly, we examine how to get the best performance in terms of sample efficiency by performing an ablation on the choices of curriculum to minimize the samples needed. This is especially important as \methodName trains two policies concurrently, which can often be unstable.
As can be seen in Figure~\ref{fig:abblation-results} (left) a strong final navigation and grasping policy can be learned with just $500$ stationary grasps while applying \textit{relabeling} to the navigation rewards. 
Next, we expand on the curriculum results on the real robot in order to tune the \textit{autonomous} curriculum to be nearly as sample efficient as the \textit{stationary} curriculum. In~\refFigure{fig:abblation-results} (middle) we can see that the best final performance is achieved by increasing the frequency of grasps early on in training and adding a bonus using the grasp model uncertainty.

\section{Discussion and Future Work}
\label{sec:discussion}

We presented \methodName, a system for autonomously learning mobile manipulation skills in the real world, without instrumentation, and with minimal human intervention. Our real-world experiments, conducted in four separate rooms, show that \methodName can train continuously for several days (40-60 hours) with only occasional interventions, and that the resulting policies are effective at cleaning up the room. Furthermore, our experiments show that \methodName continues to improve with more training, suggesting that it provides an effective approach for lifelong learning for robotic systems deployed in open-world settings. Our simulated ablation studies further provide support for the design decisions in \methodName, indicating that the hierarchical decomposition of navigation and grasping greatly improves learning performance, autonomous practicing with a suitable exploration bonus enhances learning speed, and our automated curriculum can provide effective performance when learning from scratch. Even when using the manually-provided curriculum, the grasp pretraining phase does not need to be long: only $1000$ attempted grasps were needed to get reasonable room-cleaning performance, although more pretraining can improve performance further.
Extending this system to more complex manipulation tasks would be an exciting direction for future work. We hope that this system for practical RL on mobile manipulators will make mobile manipulation more accessible to outside lab spaces and to users who need maximally autonomous learning algorithms.

\acknowledgments 

We thank the anonymous reviewers for their helpful feedback in revising our manuscript. This research was supported by the Office of Naval Research, ARL DCIST CRA W911NF-17-2-0181, the National Science Foundation, and Schmidt Futures.

\bibliography{paper}
\newpage
\appendix

\section{Algorithm Details}

\subsection{Simulated Robot}
\label{sec:simulated-robot}

Although our main experiments are conducted entirely in the real world, without any simulation, we also constructed a simulated environment to systematically evaluate and compare various algorithmic choices and ablations of our method.
We construct a simulation of our training setup using the PyBullet physics simulator. The workspace is a 3-by-3 meter room with flat walls and wooden floors. The objects consist of $20$ green spheres scattered randomly across the room without obstacles, or $30$ spheres in the room with obstacles. We evaluate our method both with and without obstacles (shown in \autoref{fig:real-robot-results} and~\ref{fig:grasp_policy_comparison} (right)). The simulated robot model is replicated to the real world and simulates how the real robot operates. During training, the environment is not reset and the same autonomous pseudo-resetting mechanism is used during training. 
During evaluation, the agent uses a greedy argmax policy to choose actions.

\subsection{Model Architectures}
\label{sec:model-architecture}
For the navigation policy, the observations are $100\times100$ RGB images from the front facing camera, which goes into three convolutional layers of sizes $64$, kernel sizes 3, and stride 1, with an average pooling layer of size 2 in between each layer. The result is passed through two dense layers of size $512$. The grasping policy uses the same observations, but center cropped to $60\times 60$ to remove unreachable regions from the image. Each of the $N=6$ networks in the grasping ensemble is also structured as three convolutional layers of size $64$, kernel size 3, and stride 2, then two dense layers of size $512$. In all experiments, we discretize the grasp action space into a $15\times15$ grid and use $\alpha=10$ and $\beta=10$ for the grasping policy, $N_{\text{grasp}}=2$, and a learning rate of $3^{-4}$ for training the networks.

\subsection{Collision Avoidance}
\label{app:collision}
To ensure that we can continue training the robot for extended periods of time, the robot must be able to safely navigate in environments with obstacles without breaking. To support safe navigation, we use the depth channel of the robot's camera to detect obstacles and walls in front of the robot. When an obstacle is detected as being inside the graspable area of the robot's arm, the robot's base rotates in a random direction until the obstacle is no longer obstructing the grasp space. We also simulate this setup for more exhaustive comparisons, with details of the simulation provided in Appendix~\ref{sec:simulated-robot}.

\subsection{Scripted Controller}
\label{sec:scripted-policy}

The scripted controller takes the camera observation, convert it to grayscale, then detects contours using OpenCV and uses the contour centroids as object locations. Then it projects the pixel locations onto the flat ground plane. For grasping, it simply grasps at the closest position. For navigation, it outputs the forward and turn amount proportional to how much it needs to reach the closest object position, clipped by the action range, or random movement if there are no object in the observation.

\section{Grasping Curriculum}
\label{sec:grasp-curriculum}
In Algorithm \ref{algo:slap_grasp_curr} and \ref{algo:slap-curriculum} we show the psuedocode of our Autonomous Curriculum algorithm. For our training, we also use hyperparameter values $N_{start} = 10$, $N_{stop} = 50$, and $N_{max} = 2000$. In addition to what was written in section $4.4$, we include an additional hyperparameters for more stable training, $N_{bt} = 300$, which determines how many grasp there needs to be in the buffer $\mathcal{D}_g$ before grasp training begins.

\begin{algorithm}[h]
\caption{TrainGraspAutoCurr($G^{1}, ..., G^{N}$, $\mathcal{D}_g$, $N$)}
\label{algo:slap_grasp_curr}
\begin{algorithmic}[1]
\If{$|\mathcal{D}_g| \geq N_{max}$}
    \State \textbf{return} TrainGrasp($G^{1}, ..., G^{N}$, $\mathcal{D}_g$, $N$, 0)
\EndIf\State{\small{\textbf{end if}}}

\State $N_{since} = 0$
\State $r_{max} = 0$

\For{$n = 1, \dots$}

    \State Get grasp observation $\Tilde{o}$.
    \State Sample $a_g \sim \pi_g(\cdot|\Tilde{o})$. // see Equation~\ref{eq:grasping-policy}
    \State Perform grasp $a_g$, receiving $r_g = 0$ or $1$.
    \State Store $(\Tilde{o}, a_g, r_g)$ in $\mathcal{D}_g$.
    
    \If{$|\mathcal{D}_g| \geq N_{bt}$}
        \State Update $G^{1}, ..., G^{N}$ on $\mathcal{D}_g$
    \EndIf\State{\small{\textbf{end if}}}
    
    \If{$r_g = 1$}
        \State $N_{since} = 0$
        \State $r_{max} = 1$
    \Else
        \State $N_{since} = N_{since} + 1$
    \EndIf\State{\small{\textbf{end if}}}
    
    \If{$|\mathcal{D}_g| \geq N_{max}$}
        \State \textbf{break}
    \EndIf\State{\small{\textbf{end if}}}
    
    \If{$r_{max} = 1$}
        \If {$N_{since} \geq N_{stop}$}
            \State \textbf{break}
        \EndIf\State{\small{\textbf{end if}}}
    \Else
        \If {$N_{since} \geq N_{start}$}:
            \State \textbf{break}
        \EndIf\State{\small{\textbf{end if}}}
    \EndIf\State{\small{\textbf{end if}}}
    
    \If {$r_g = 1$}:
        \State Drop object randomly in grasping area.
    \EndIf\State{\small{\textbf{end if}}}
\EndFor\State{\small{\textbf{end for}}}
\State \textbf{return} $r_{max}$
\end{algorithmic}
\end{algorithm}

\begin{algorithm}[h]
\caption{\methodname~with AutoCurr}
\label{algo:slap-curriculum}
\begin{algorithmic}[1]
\State {Init: function estimators $\pi_n$, $G^{1}, \ldots, G^{M}$}.
\State{Replay buffers $\mathcal{D}_n = \{\}$, $\mathcal{D}_g = \{\}$}
\State \changes{Deleted TrainGrasp($G^{1}, .., G^{N}$, $D_g$, $N_\text{pt}$, 1)}
\For{$t = $ 0, \dots, $T$ steps}
    \State Get navigation observation $o_t$
    \State Sample $a_n \sim \pi_n(\cdot|o_t)$ and perform $a_n$
    \If{ \textit{uniform}() $\leq$ $\mathbb{P}[\text{grasp}|o_t]$}
        \State {$r_{g}\!=$\changes{TrainGraspAutoCurr($G^{1},\! .., G^{M}$,$D_g$,$N_\text{grasp}$)} \label{alg:slap-grasp-train}}
    \Else { $r_g = 0$}
    \EndIf
    
    \State Compute reward $r_n = r_g - 1$
    \State Get next navigation observation $o_{t+1}$
    \State Store $(o_t, a_n, r_n, o_{t+1})$ in $\mathcal{D}_n$.
    \State Update $\pi_n$ with $\mathcal{D}_n$ using SAC.
    \State Pseudo-reset
\EndFor
\State {\small{\textbf{end for}}}
\end{algorithmic}
\end{algorithm}

\section{Additional Tasks}
\label{sec:pick-and-place}

\begin{figure}[b]
    \centering
    \includegraphics[width=0.35\linewidth]{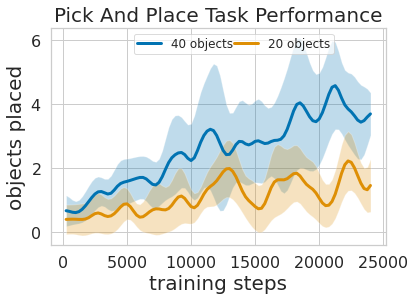}
    \includegraphics[width=0.3\linewidth]{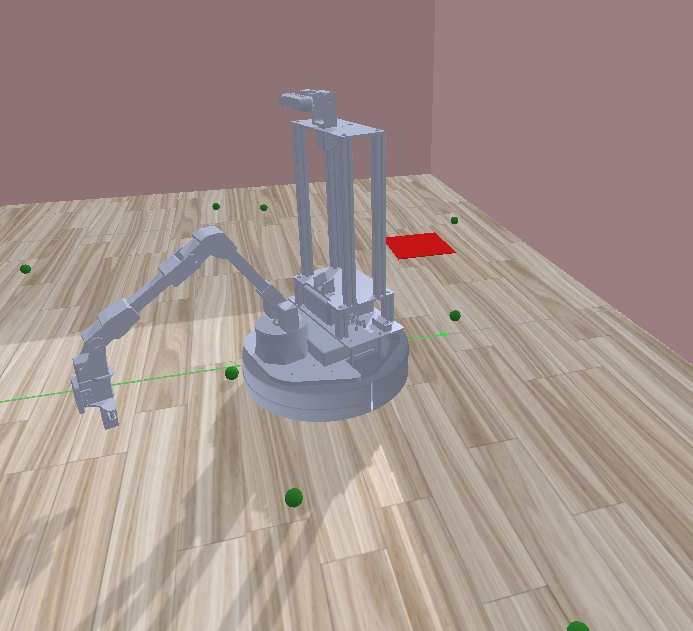}
    \caption{
    Left: Results for pick and place training, with either 40 or 20 objects available in the environment. We allow a maximum of 250 simulation steps for each evaluation rollout.
    Right: A snapshot from the proposed pick and place task with the red pace target behind the robot.
    }
    \label{fig:pick_and_place}
\vspace{-0.5cm}
\end{figure}

Our main experiments use the room cleanup task, but the basic design of the ReLMM system can also be extended to other tasks, as we discuss in this section. Using the simulation environment, we evaluate ReLMM on a picking and placing task, which requires picking up objects and placing them on a red rectangle (see Figure~\ref{fig:pick_and_place}). The policy is trained in a similar fashion as the grasping policy, with a sparse reward, and uses the same action space. It is given reward of 1 if the object is successfully placed on one of the designated areas marked red, shown in \autoref{fig:pick_and_place}, and 0 otherwise. During training, we strictly follow the Algorithm \ref{algo:slap} and sample the actions according to the distribution defined in Equation \ref{eq:grasping-policy}. In step 9 of the algorithm the decision of whether to use the grasping or placing policy depends on whether or not an object is already held by the end effector. In contrast to the grasping task, the placing task does not use the ReLMM pseudo-reset mechanism, and instead simply learns to place objects down at random positions in the environment. Analogously to the pretraining procedure we use for the grasping task, we pretrain the grasping and placing policies with $N_\text{grasp,st}$, $N_\text{place,st} = 2000$ samples, which matches the number used for pretraining the original grasping task. $N_\text{obj}$ is the number of objects available for grasping ($20$ and $40$), and we use $N_\text{targets} = 3$ random targets for placing.
Since the placing portion of the task is easier compared to the more demanding grasping task, we achieve ~71\% success rate during placing pretraining compared to ~60\% for the grasping. During the combined training process, the navigation policy is given a reward -1 when not holding an object, -0.5 if the robot is holding the successfully grasped object, and a reward of 0 when the object is successfully placed on a red target, which also terminates the episode. The positions of the target areas are randomized after each successful place. This experiment illustrates that the ReLMM system is general enough to be adapted to other tasks with more complex reward structures.

\section{Additional Results}
\label{app:results}

Here we include additional results for the paper that include images from the real robot experiments and add to the ablation analysis.

\subsection{Additional Simulation Results}

\begin{figure}
    \centering
    \includegraphics[width=0.3\linewidth]{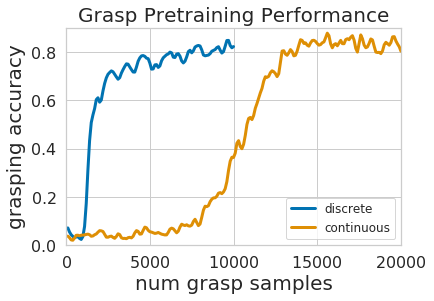}
    \includegraphics[width=0.3\linewidth]{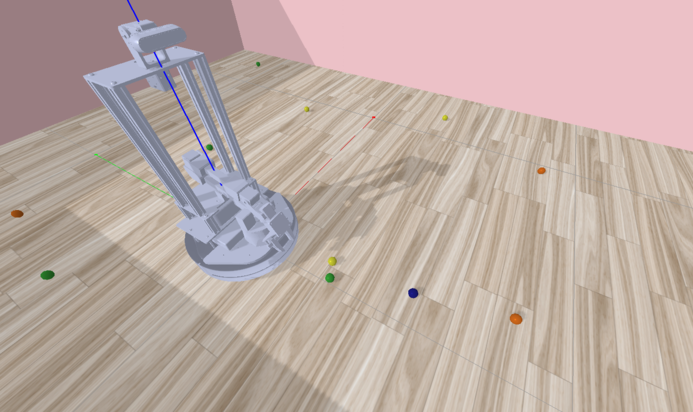}
    \caption{
    Left: Discrete grasping policies train significantly faster than continuous policies in our simulation ablation study.
    Right: Images from the simulated environment with diverse objects.
    }
    \label{fig:grasp_policy_comparison}
\vspace{-0.5cm}
\end{figure}

\begin{figure}
    \centering
    \includegraphics[width=0.45\linewidth]{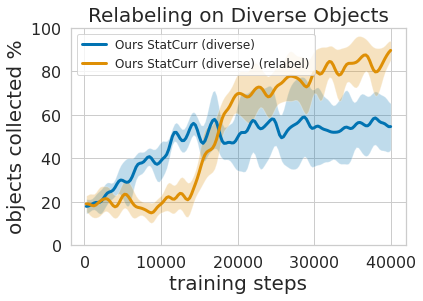}
    \includegraphics[width=0.45\linewidth]{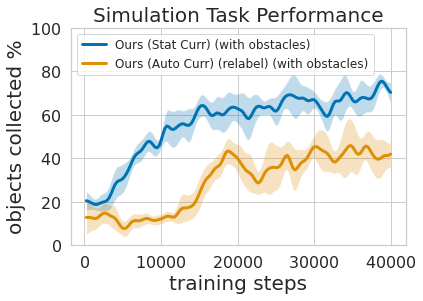}
    \caption{
    Further curriculum and relabeling comparison in simulation. We plot the performance for the simulated room with diverse objects (left) and with obstacles (right). We find that the relabeling reward helps significantly with the diverse objects because it encourages the navigation to go towards areas of high grasp uncertainty. However, the automatic curriculum in the obstacle room is still slower than the stationary curriculum, even with the best hyperparameters.}
    \label{fig:more-curriculum}
\vspace{-0.5cm}
\end{figure}

In simulation we study three additional conditions for \methodName. Starting with a comparison of using a \textit{discrete} vs \textit{continuous} action representation. In this experiment the action space uses the same underlying control structure of finding the best X - Y position to grasp on the ground, only the output policy distribution changes. The results in~\autoref{fig:grasp_policy_comparison} show that training with a discrete policy trains more than twice as fast. 
Next, we study the affect of using the learned grasping model to \textit{relabel} rewards for the navigation policy, as described in~\autoref{par:nav}. We find that using this relabeling method can increase learning speed and final policy quality, as shown in~\autoref{fig:more-curriculum} (left).
Last, we compare the different curriculum methods, stationary and autonomous, in a simulated room with obstacles. The stationary curriculum method appears to results in faster learning but requires human intervention to train. The autonomous curriculum may find this environment difficult do to the obstacles which make exploratory navigation less successful.

\paragraph{Ablation.} We compare our pseudo-reset method to a prior algorithm for reset-free learning~\cite{Zhu2020}, which learns how to perturb the environment between episodes of running the actual task policy by maximizing a novelty based reward. In~\autoref{fig:abblation-results} (right) we see our simpler pseudo-reset is equally or more effective than the learned perturbation method, without requiring any additional learning machinery.

\subsection{Comparison to HRL4IN}

For HRL4IN \cite{li2020hrl4in}, we followed the method laid out in the HRL4IN paper. To make the setup match the one in our experiments, the observation space contains the RGB camera image (instead of the depth image HRL4IN uses), global XYZ position of the robot (which is not available on the real robot and not used by our method), and the local position of the gripper. The action spaces for the high-level and low-level are the velocity of the wheels and the change in gripper XYZ position. In the same manner as ReLMM, when the gripper’s height above ground goes below some threshold, the gripper closes and picks from the ground. We use the same high-level policy reward that we employ in ReLMM, where the low-level policy uses the same as the HRL4IN paper. After around $30$ days of training in simulation, HRL4IN has collected a total of 30M environment steps, but the performance is still at around 1\% objects collected on average during eval. This level of performance is expected, since the HRL4IN paper also took around 30M environment steps to achieve good performance on their task, but our environment is a sparse reward setting, whereas the HRL4IN paper uses a dense reward environment. On the other hand, our method only required around 30K environment steps to learn a strong policy with 90\% objects collected on average during eval. Our method is focused on real-world training, and significantly more efficient.

\end{document}